\pdfoutput=1

\documentclass[11pt]{article}

\usepackage[final]{acl}
\usepackage{times}
\usepackage{latexsym}
\usepackage[T1]{fontenc}
\usepackage[utf8]{inputenc}
\usepackage{microtype}
\usepackage{inconsolata}
\usepackage{graphicx}
\usepackage{amsmath, amssymb}
\usepackage{algorithm}
\usepackage{algpseudocode}
\usepackage{booktabs}
\usepackage{multirow}
\usepackage{url}
\usepackage{float}
\usepackage{tabularx}
\usepackage{enumitem}

\title{DFPE: A Diverse Fingerprint Ensemble for Enhancing LLM Performance
}

\author{
  Seffi Cohen \\
  \And
  Niv Goldshlager \\
  \And
  Nurit Cohen-Inger \\
  Ben Gurion University, Beer Sheva, 8410501, Israel
  \And
  Bracha Shapira \\
  \And
  Lior Rokach \\
}
\date{}

\begin{document}
\maketitle

\begin{abstract}
Large Language Models (LLMs) have shown remarkable capabilities across various natural language processing tasks but often struggle to excel uniformly in diverse or complex domains. We propose a novel ensemble method - Diverse Fingerprint Ensemble (DFPE), which leverages the complementary strengths of multiple LLMs to achieve more robust performance. Our approach involves: (1) clustering models based on response "fingerprints" patterns, (2) applying  a quantile-based filtering mechanism to remove underperforming models at a per-subject level, and (3) assigning adaptive weights to remaining models based on their subject-wise validation accuracy. In experiments on the Massive Multitask Language Understanding (MMLU) benchmark, DFPE outperforms the best single model by 3\% overall accuracy and 5\% in discipline-level accuracy. This method increases the robustness and generalization of LLMs and underscores how model selection, diversity preservation, and performance-driven weighting can effectively address challenging, multi-faceted language understanding tasks.
\end{abstract}

\section{Introduction}

Large Language Models (LLMs) have demonstrated remarkable capabilities across a wide range of natural language processing tasks ~\cite{chang2024survey, matarazzo2025survey}. Yet, when confronted with complex multitask benchmarks such as the Massive Multitask Language Understanding (MMLU)~\cite{hendrycks2020measuring}, a single LLM often struggles to excel uniformly across all subjects. The MMLU benchmark encompasses diverse subjects, each varying in difficulty and domain breadth. These variations pose significant challenges, revealing performance gaps that no individual model can easily bridge.

Ensembling multiple LLMs provides a promising avenue to overcome these limitations \cite{tekin2024llm, jiang2023llm, mavromatis2024pack, xu2025hit}. By leveraging the complementary strengths of different models, an ensemble can achieve greater accuracy, robustness, and adaptability than any single model alone \cite{lu2024merge}. However, maximizing the potential of an ensemble requires careful selection and integration of its models. Key challenges include identifying models that offer complementary strengths, preserving a diverse set of solution strategies, adapting to subject-specific difficulties, and efficiently aggregating predictions.
\begin{table*}[h!]
\centering
\caption{Comparison of related methods. ``Zero/Few-Shot Training'' indicates methods that do not require additional training or fine-tuning. DFPE requires no additional training, uses few-shot validation for subject adaptivity, and ensures diversity via clustering and quantile-based filtering.}
\begin{tabular}{p{2.9cm} c c c}
\toprule
\textbf{Method} & \textbf{Zero/Few-Shot Training} & \textbf{Subject Adaptivity} & \textbf{Diversity Optimization} \\
\midrule
LLM-TOPLA & \texttimes  & \checkmark & \checkmark \\
SelectLLM & \texttimes  & \checkmark & \texttimes \\
PackLLM & \texttimes & \checkmark & \checkmark \\
SweetSpan & \checkmark  & \texttimes & \checkmark \\
DeePEn & \texttimes  & \checkmark & \checkmark \\
EVA & \checkmark  & \texttimes & \checkmark \\
Boosted Prompts & \checkmark  & \texttimes & \checkmark \\
CAPE & \checkmark  & \checkmark & \texttimes \\
ZOOTER & \texttimes  & \checkmark & \checkmark \\
LoRA Ensembles & \texttimes  & \texttimes & \checkmark \\
\textbf{DFPE (Ours)} & \checkmark  & \checkmark & \checkmark \\
\bottomrule
\end{tabular}
\label{tab:related_work_comparison}
\end{table*}
In this paper, we present the Diverse Fingerprint Ensemble (DFPE) method that aims to optimize the ensemble for a particular subject, thereby improving the performance of LLMs on that subject. Our approach optimizes the ensemble of models by combining effective model selection, subject-level adaptivity, and adaptive weighting. The key contributions of our method include:
\begin{itemize}[leftmargin=*,noitemsep]
    \item \textbf{Diversity Preservation through Response Pattern Clustering}: To ensure the ensemble incorporates diverse problem-solving strategies, we capture each model's response patterns through validation-set "fingerprints" and cluster them using DBSCAN. This systematically maintains complementary approaches while filtering redundancy, preventing the ensemble from converging to a single solution path and enhancing its ability to handle varying question types.
    
    \item \textbf{Subject-Specialized Expertise Allocation}: To balance domain-specific capabilities with strategic diversity, we implement a two-stage process: first clustering models with similar response patterns, then selecting the highest-performing model from each cluster based on subject-specific validation accuracy. These selected models receive weights scaled by their subject performance, creating an ensemble that combines diverse approaches while emphasizing proven expertise.
    
    \item \textbf{Adaptive Performance Thresholding}: To maintain quality while preserving valuable minority perspectives, we employ subject-level quantile-based filtering that automatically adjusts to varying difficulty levels across disciplines. This approach ensures high standards while accommodating the inherent differences between subjects, retaining models that might excel on specific question types.
\end{itemize}

We evaluate our approach on the MMLU benchmark using a diverse pool of up to 9 billion parameters LLMs. Our ensemble achieves a final accuracy of 73.5\%, outperforming the best single model by 3\%, and improves the discipline-level accuracy by 5\%. These results demonstrate that our approach, which prioritizes selecting the most diverse and accurate LLMs tailored to each subject, leads to significant improvements in performance on complex, multi-faceted language understanding tasks.

\begin{figure*}[h!]
\centering
\includegraphics[width=\linewidth]{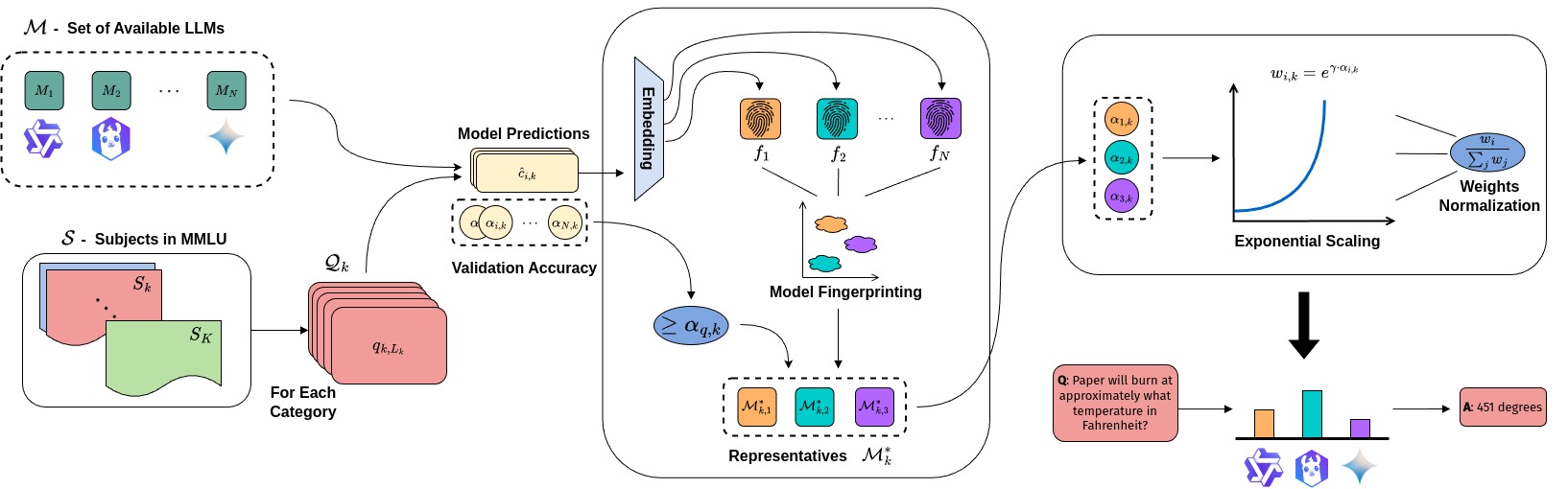}
    \caption{A pool of LLMs \(\mathcal{M}\) is evaluated per subject \(S_k \in \mathcal{S}\) using a small validation set \(\mathcal{Q}_k\). 
    Each model’s predictions and accuracy \(\alpha_{i,k}\) are used to generate “fingerprints,” which are subsequently clustered to maintain diversity. 
    Models failing to meet a subject-specific quantile threshold are removed, and the most accurate model is chosen from each cluster to form the representative set \(\mathcal{M}_k^*\). 
    An exponential weighting scheme is then applied to these representatives before their final, weighted votes are aggregated to produce the answer}
\label{fig:dfpe_pipeline}
\end{figure*}
\section{Related Work}

Ensembling techniques have been widely explored as a means to boost the performance of LLMs on complex multitask benchmarks like MMLU. These methods typically aim to balance improved accuracy with practical constraints such as computational overhead, training requirements, and the diversity of model contributions. Some strategies focus on maximizing diversity across model outputs to exploit complementary strengths, while others emphasize efficient routing or fine-tuning to adapt ensemble components.

A prominent direction involves assembling multiple models to capitalize on their unique abilities, as seen in LLM-TOPLA~\cite{tekin2024llm}, PackLLM~\cite{mavromatis2024pack}, and SweetSpan~\cite{xu2025hit}, each of which coordinates output information at various granularities (e.g., token- or span-level). While these techniques often produce strong results, they may also introduce extra training or inference overhead. Similar concepts appear in ZOOTER~\cite{lu2023routing}, which dynamically routes queries to suitable models, and LoRA Ensembles~\cite{wang2023lora}, which uses parameter-efficient fine-tuning to merge model capabilities. However, approaches that rely heavily on additional training or fine-tuning can be less practical when resources are limited or when rapid experimentation is required.

Other methods seek to remain fully or mostly training-free. Boosted Prompts~\cite{pitis2023boosted} and CAPE~\cite{jiang2023calibrating}, for instance, forgo further training in favor of leveraging pretrained models and calibrating their outputs. This philosophy aligns with EVA~\cite{xu2024bridging} and DeePEn~\cite{huang2024ensemble} in that they preserve model diversity or transform model outputs at an inference or validation stage without extensive fine-tuning. However, many of these solutions do not adapt well to subject-wise performance variations, which becomes crucial for wide-ranging benchmarks like MMLU.

SelectLLM~\cite{maurya2024selectllm} provides an example of per-query adaptivity, where a routing scheme decides which model should handle a particular input. Yet, these methods may not systematically ensure the retention of diverse solution paths or fine-grained subject-level adaptivity.

Table~\ref{tab:related_work_comparison} compares these ensemble approaches with respect to (a) zero/few-shot training needs, (b) subject adaptivity, and (c) explicit diversity optimization. As shown, many existing solutions either require training or lack robust subject-wise specialization. Our method, DFPE, is distinguished by its reliance on few-shot validation rather than intensive training, its explicit per-subject clustering and filtering to maintain competence and diversity, and an adaptive weighting mechanism that emphasizes strong models without discarding minor yet potentially useful perspectives. This combination ensures that DFPE remains both robust and subject adaptive, providing a substantial improvement on language understanding tasks without demanding additional fine-tuning.

Overall, DFPE stands out by forgoing further training or complex fine-tuning, relying instead on validation-driven selection and weighting to construct a subject-adaptive, diversity-optimized ensemble that excels on challenging multitask benchmarks like MMLU.

\section{Method}
\label{sec:method}

In this section, we detail our ensemble method designed to optimize LLMs ensemble for a specific subject tasks. Our approach integrates clustering-based selection, quantile-based filtering, and adaptive weighting to ensure that only sufficiently capable and diverse models contribute to the final ensemble prediction. By leveraging per-subject validation performance and subject-specific fingerprints, we construct a more balanced and effective ensemble than naive averaging or manual heuristics.

\subsection{Overview}
As illustrated in Figure~\ref{fig:dfpe_pipeline}, our method involves four main steps:
\begin{enumerate}[leftmargin=*,noitemsep]
    \item \textbf{Model Fingerprinting and Clustering}: Represent each model’s responses on a subject as a fingerprint vector. Cluster these fingerprints using DBSCAN with cosine similarity, grouping models exhibiting similar response patterns.
    \item \textbf{Quantile-based Model Selection}: Identify a quantile-based accuracy threshold per subject and retain only models meeting or exceeding this benchmark. From each cluster, select the model with the highest accuracy, to ensure both competence and diversity.
    \item \textbf{Adaptive Weighting Scheme}: Assign weights to the selected models using exponential scaling of their per-subject validation accuracy. Normalize these weights to form a proper distribution.
    \item \textbf{Ensemble Prediction}: Aggregate the weighted predictions of the selected models via a weighted voting mechanism, producing the final answer.
\end{enumerate}

This pipeline ensures that the ensemble benefits from both model diversity and subject-specific adaptivity, with performance-driven weighting giving more influence to stronger models.

\subsection{Formal Description}

\paragraph{Notation}

\begin{itemize}[leftmargin=*,noitemsep]
    \item Let $\mathcal{M} = \{M_1, M_2, \dots, M_N\}$ be the set of available LLMs.
    \item Let $\mathcal{S} = \{S_1, S_2, \dots, S_K\}$ represent the subjects in MMLU.
    \item For each subject $S_k$, let $\mathcal{Q}_k = \{q_{k,1}, q_{k,2}, \dots, q_{k,L_k}\}$ denote its questions.
    \item Each question $q_{k,l}$ has choices $\mathcal{C}_{k,l} = \{c_{k,l,1}, \dots, c_{k,l,C_{k,l}}\}$.
    \item Let $\hat{c}_{i,k,l}$ be the prediction of model $M_i$ for question $q_{k,l}$.
    \item Let $\alpha_{i,k}$ denote the validation accuracy of model $M_i$ on subject $S_k$.
\end{itemize}

\paragraph{Model Fingerprinting and Clustering}

For each model $M_i$ and subject $S_k$, we produce a fingerprint vector $\mathbf{f}_{i,k}$ summarizing the model’s response behavior on validation data. We use the pre-trained sentence embedding model all-MiniLM-L6-v2 \cite{huggingface2023sentence} to embed each output and average the embeddings to create the fingerprint.. Once we obtain $\{\mathbf{f}_{i,k}\}_{i=1}^N$, we apply DBSCAN \cite{ester1996density} clustering with cosine similarity to group models into clusters based on their response patterns.

\paragraph{Quantile-based Model Selection}

Define a quantile parameter $q$ (e.g., $q=0.1$) to set a subject-specific accuracy threshold. Let $\{\alpha_{i,k}\}_{i=1}^N$ be the accuracies of all models on $S_k$. Compute the $q$-quantile accuracy threshold  $\alpha_{q,k}$ for subject $S_k$. Retain models satisfying $\alpha_{i,k} \geq \alpha_{q,k}$, ensuring a baseline level of competence per subject.

From the filtered models, select the representative model from each cluster:
\[
M^*_{k,j} = \arg\max_{M_i \in C_j} \alpha_{i,k}.
\]
The set of all such representatives for subject $S_k$ is $\mathcal{M}^*_k$.

\paragraph{Adaptive Weighting Scheme}

For each selected model $M_i \in \mathcal{M}^*_k$, we assign a weight:
\[
w_{i,k} = \exp(\gamma \cdot \alpha_{i,k}),
\]
where $\gamma$ is a scaling factor that highlights performance differences. Normalize these weights so that they sum to 1:
\[
\tilde{w}_{i,k} = \frac{w_{i,k}}{\sum_{M_j \in \mathcal{M}^*_k} w_{j,k}}.
\]

\paragraph{Ensemble Prediction}

For a test question $q_{k,l}$ in subject $S_k$, each selected model $M_i \in \mathcal{M}^*_k$ provides a predicted choice $\hat{c}_{i,k,l}$. We aggregate their votes using the assigned weights:
\[
V_{k,l}(c) = \sum_{M_i \in \mathcal{M}^*_k} \tilde{w}_{i,k} \cdot \mathbb{I}[\hat{c}_{i,k,l}=c].
\]
The final ensemble prediction is:
\[
\hat{c}_{k,l} = \arg\max_{c \in \mathcal{C}_{k,l}} V_{k,l}(c).
\]

\begin{algorithm}[H]
\caption{DFPE (Diverse Fingerprint Ensemble)}
\begin{algorithmic}
\For{each subject $S_k$}
   \State \textbf{1.} For models $M_i$, compute accuracy $\alpha_{i,k}$ on validation $\mathcal{Q}_k$, filter out those below $q$-quantile.
   \State \textbf{2.} Create fingerprints $\mathbf{f}_{i,k}$ (e.g., embeddings), cluster via DBSCAN.
   \State \textbf{3.} From each cluster, pick the model $M_i$ with highest $\alpha_{i,k}$; call this set $\mathcal{M}_k^*$.
   \State \textbf{4.} Assign weights $w_{i,k} = \exp(\gamma \alpha_{i,k})$ to models in $\mathcal{M}_k^*$ and normalize.
\EndFor
\For{each test question $q_{k,l}$ in $S_k$}
   \State Aggregate predictions with weighted votes:
   \[
   \hat{c}_{k,l}=\arg\max_{c}\sum_{M_i\in\mathcal{M}_k^*} \bigl(w_{i,k}\cdot\mathbb{I}[\hat{c}_{i,k,l}=c]\bigr).
   \]
\EndFor
\end{algorithmic}
\end{algorithm}

\subsection{Rationale}

Our method strikes a balance between diversity, competence, and adaptability:
\begin{itemize}[leftmargin=*,noitemsep]
    \item \textbf{Diversity Preservation}: Clustering models ensures variety in solution strategies, reducing redundancy.
    \item \textbf{Quantile-based Competence}: The quantile threshold removes weak models, guaranteeing a baseline performance level.
    \item \textbf{Adaptive Weighting}: Higher accuracy models naturally exert greater influence, but all retained models still contribute to the final decision.
    \item \textbf{Practical Simplicity}: Operating at the output level avoids complex token- or span-level computations, making the method more scalable.
\end{itemize}

By combining these elements, our approach effectively exploits the complementary strengths of multiple LLMs, yielding improved performance on complex multitask language understanding.

\section{Experimental Setup}
\label{sec:experimental_setup}

In this section, we detail the dataset, base models, and implementation specifics used to evaluate our improved ensemble method. We rely on the Massive Multitask Language Understanding (MMLU) benchmark~\cite{hendrycks2020measuring}, incorporate a set of Large Language Models (LLMs) with varying capabilities, and use few-shot validation to guide quantile-based selection and adaptive weighting. We present the final results alongside attached figures illustrating discipline-level performance, sensitivity analyses, and accuracy comparisons.

\subsection{Dataset}

We evaluate our method on the MMLU benchmark~\cite{hendrycks2020measuring}, which consists of 57 subjects spanning a diverse range of disciplines, including STEM fields, humanities, and social sciences. Each subject comprises multiple-choice questions in English, which vary in difficulty. The given benchmark data is partitioned into a few-shot validation set for model selection and fingerprint generation, and a separate test set for the final evaluation. Overall, there are 14,079 test samples and 1,540 validation samples.
The MMLU dataset is distributed under the MIT License, which allows for free use, modification, and distribution as long as the original copyright notice and license terms are maintained.

\begin{table}[H]
\centering
\caption{Overall accuracies of the base LLMs on the MMLU benchmark. The diverse range of performance provides a strong foundation for our quantile-based ensemble approach.}
\begin{tabular}{l c c}
\toprule
\textbf{Model} & \textbf{Params} & \textbf{Accuracy} \\
\midrule
GLM-4 & 9B & 0.7076 \\
Qwen2.5 & 7B & 0.6476 \\
Qwen2.5 & 3B & 0.6605 \\
Mistral v0.3 & 7B & 0.6316 \\
Phi-3.5-mini & 4B & 0.7046 \\
Llama-3.1 & 8B & 0.6508 \\
Gemma-2 & 9B & 0.6804 \\
Apollo2 & 7B & 0.7034 \\
Starling-LM-alpha & 7B & 0.6149 \\
Yi-1.5 & 6B & 0.6239 \\
\bottomrule
\end{tabular}
\label{tab:base_models}
\end{table}

\subsection{Base Models}

Our ensemble is constructed from a pool of up to 9B parameters open LLMs differing in architecture, size, and training methodology, including GLM-4-9B-chat \cite{glm2024chatglm}, Qwen2.5-7B-Instruct \cite{yang2024qwen2}, Qwen2.5-3B-Instruct \cite{yang2024qwen2}, Mistral-7B-Instruct-v0.3 \cite{jiang2023mistral}, Phi-3.5-mini-instruct \cite{abdin2024phi}, Llama-3.1-8B-Instruct \cite{dubey2024llama}, Gemma-2-9B \cite{gemma_2024}, Apollo2-7B \cite{zheng2024efficiently}, Starling-LM-7B-alpha \cite{starling2023}, and Yi-1.5-6B \cite{young2024yi}. These models are detailed in Table \ref{tab:base_models}. The best single model in our experiments achieves 70.76\% overall-accuracy on MMLU, with a discipline-accuracy of 69\%. Other models range from approximately 61.5\% to 70.5\% overall-accuracy, providing a rich diversity of performance levels. This variability creates an opportunity to exploit complementary strengths: by carefully selecting and weighting models via quantile-based filtering and clustering, we aim to surpass the capabilities of any single model.

\subsection{Implementation Details}
\paragraph{Reproducible Code}
The source code and raw experiment results are available at \url{https://github.com/nivgold/DFPE}.
\paragraph{Clustering and Quantile Thresholding:}  
For each subject, we embed each model’s validation responses to form a fingerprint vector. We then cluster these fingerprints using DBSCAN with cosine similarity. The DBSCAN epsilon parameter is chosen based on empirical tuning to ensure stable, meaningful clusters.  
We select a quantile parameter $q$ = 0.05 based on initial validation performance. Models meeting or exceeding the $q$-quantile accuracy threshold are retained.
\paragraph{Adaptive Weighting and Hyperparameters:}  
We apply exponential scaling to each retained model’s per-subject validation accuracy. Parameters such as the AccuracyFactor and quantile thresholds are determined via validation-based exploration. For example, setting the AccuracyFactor (\(\gamma\)) to 5.0 and choosing $q=0.05$ consistently yields strong performance.
\paragraph{Few-Shot Validation for Selection and Weighting:}
For each subject, given validation questions (approximately 10\% of the test set) to measure model accuracy. These validation accuracies guide the quantile threshold, clustering-based selection, and the adaptive weighting scheme. 
\paragraph{Hardware Details:}
All experiments were conducted on RTX 6000 GPU, which provided sufficient computational resources for up to 9B parameters LLMs.

\subsection{Evaluation Metrics}

We report both the Overall Accuracy and Discipline-Accuracy which accounts for discipline-level balance, ensuring improvements are not limited to a few disciplines. We compare four configurations:
\begin{itemize}[noitemsep,leftmargin=*]
    \item \textbf{Best Single Model (BSM)}: The single model achieving the highest test accuracy is which accounts for subject-level balance, ensuring improvements are not limited to a few subjects.
    \item \textbf{Best Single Model on Validation (BSMoV)}: The model that performs best on the validation set.
    \item \textbf{Majority Voting Ensemble (MVoting)}: Equal weights ensemble method.
    \item \textbf{Our Ensemble (DFPE)}: The proposed quantile-driven cluster-based selection method with adaptive weighting is named DFPE.

\end{itemize}

These comparisons demonstrate how DFPE can outperform all the compared methods.

\section{Results}
\label{sec:results}
We evaluate the performance of DFPE through extensive experiments, analyzing overall accuracy improvements, discipline-specific gains, and the impact of various hyperparameters. To enhance clarity, we aggregated the results at the subject level into discipline-level outcomes. Our findings demonstrate significant and consistent improvements over baseline approaches across a wide range of disciplines.

\subsection{Overall Performance}
DFPE consistently outperforms all comparison methods on the MMLU benchmark. As shown in Table~\ref{tab:accuracy_results}, it achieves higher overall accuracy and discipline-specific accuracy than the best single models (BSM, BSMoV) as well as majority voting (MVoting). These results underscore the effectiveness of DFPE’s design, which leverages quantile-based filtering to identify top-performing predictions, clustering to preserve diverse model insights, and adaptive weighting to integrate complementary strengths into a unified ensemble.

\begin{table}[h]
\centering
\caption{Accuracy and Discipline-Accuracy Comparison.
DFPE significantly outperforms the Best
Single Model (BSM), the best Single Model on Validation (BSMoV), and the Majority Voting method (MVoting).}
\resizebox{0.5\textwidth}{!}{%
\begin{tabular}{lcc}
\hline
\textbf{Model} & \textbf{Accuracy} & \textbf{Discipline-Accuracy} \\
\hline
BSM & 0.708  & 0.690 \\
BSMoV & 0.677 & 0.676 \\
MVoting & 0.724  & 0.727 \\
\textbf{DFPE (Optimal)} & \textbf{0.735} & \textbf{0.740} \\

\hline
\end{tabular}
}
\label{tab:accuracy_results}
\end{table}

\subsection{Discipline-level Analysis} 
\begin{figure*}[h!]
\centering
\includegraphics[width=\linewidth]{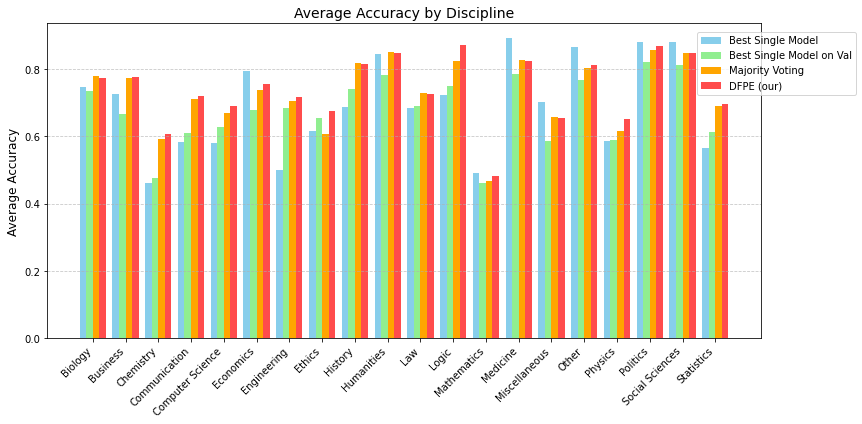}
\caption{Average Accuracy by Discipline. DFPE consistently outperforms the compared methods across a wide range of disciplines, highlighting broad-spectrum gains.}
\label{fig:discipline_performance}
\end{figure*}

Table~\ref{tab:subject_results} provides a detailed breakdown of our results across 20 distinct disciplines. Overall, DFPE attains the highest average accuracy (0.740) and achieves the top score in 9 out of 20 disciplines (Business, Chemistry, Communication, Computer Science, Engineering, Ethics, Logic, Physics, and Statistics). 
As visualized in Figure~\ref{fig:discipline_performance}, DFPE consistently matches or outperforms competing methods across a diverse set of tasks. These results underscore the strength of DFPE’s diversity-preserving design, which allows it to integrate the complementary knowledge of multiple base models and deliver robust, broad-spectrum performance.

\begin{table}[h]
\centering
\caption{Comparison of discipline-level Results}
\resizebox{0.5\textwidth}{!}{%
\begin{tabular}{lcccc}
\hline
\textbf{Discipline} & \textbf{BSM} & \textbf{BSMoV} & \textbf{MVoting} & \textbf{DFPE} \\
\hline
Biology           & 0.747 & 0.734 & \textbf{0.777} & 0.773 \\
Business          & 0.724 & 0.665 & 0.773 & \textbf{0.775} \\
Chemistry         & 0.460 & 0.475 & 0.593 & \textbf{0.605} \\
Com               & 0.583 & 0.609 & 0.709 & \textbf{0.718} \\
Comp Sci          & 0.581 & 0.629 & 0.668 & \textbf{0.690} \\
Economics         & \textbf{0.794} & 0.678 & 0.738 & 0.756 \\
Engineering       & 0.500 & 0.683 & 0.703 & \textbf{0.717} \\
Ethics            & 0.614 & 0.654 & 0.607 & \textbf{0.674} \\
History           & 0.686 & 0.741 & \textbf{0.818} & 0.815 \\
Humanities        & 0.844 & 0.781 & \textbf{0.851} & 0.847 \\
Law               & 0.683 & 0.690 & \textbf{0.729} & 0.726 \\
Logic             & 0.722 & 0.749 & 0.822 & \textbf{0.871} \\
Math              & \textbf{0.489} & 0.462 & 0.468 & 0.481 \\
Medicine          & \textbf{0.890} & 0.784 & 0.825 & 0.823 \\
Misc              & \textbf{0.701} & 0.586 & 0.656 & 0.654 \\
Other             & \textbf{0.864} & 0.766 & 0.804 & 0.810 \\
Physics           & 0.585 & 0.590 & 0.615 & \textbf{0.650} \\
Politics          & \textbf{0.878} & 0.822 & 0.856 & 0.868 \\
Social Sci        & \textbf{0.880} & 0.810 & 0.847 & 0.848 \\
Statistics        & 0.565 & 0.611 & 0.690 & \textbf{0.694} \\
\hline
\textbf{Average}  & 0.690 & 0.676 & 0.727 & \textbf{0.740} \\
\hline
\end{tabular}
}
\label{tab:subject_results}
\end{table}

\subsection{Sensitivity Analysis}
\begin{figure*}[h]
\centering
\includegraphics[width=\linewidth]{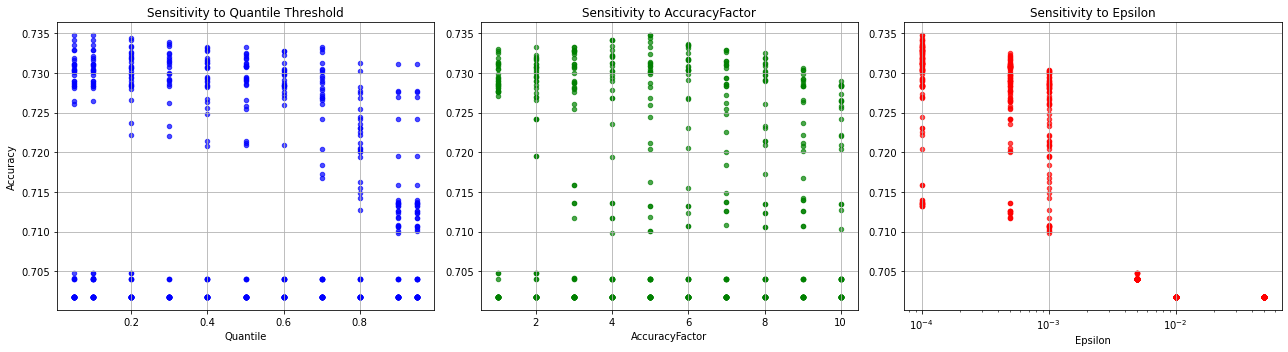}
\caption{Sensitivity analysis. Left: Accuracy vs. Quantile Threshold; Middle: Accuracy vs. AccuracyFactor; Right: Accuracy vs. Epsilon - the Epsilon axis (log scale). Performance remains robust within moderate parameter ranges, easing the tuning process.}
\label{fig:sensitivity_analysis}
\end{figure*}

We conduct comprehensive sensitivity analyses across three key hyperparameters, as visualized in Figure~\ref{fig:sensitivity_analysis}:

\textbf{Quantile Threshold}: Optimal performance at 0.05. \\
- Stable performance range: 0.05-0.2 \\
- Performance degradation beyond 0.10

\textbf{AccuracyFactor}: Best results at 5. \\
- Effective range: 4.0-6.0 \\
- Diminishing returns beyond 6.0

\textbf{DBSCAN Epsilon}: Robust clustering achieved with low epsilon. \\
- Stable performance: 0.0001-0.0005 \\
- Higher epsilon values result in an insufficient number of clusters, and when the epsilon value exceeds 0.01, all models fall into the same cluster, yielding results equivalent to the best single model.

\subsection{Model Participation and Efficiency}
In our optimal accuracy configuration, DFPE typically retains most of the available models (9 out of 10) per subject. This suggests that in the highest-performing setting, the diversity and complementary strengths of nearly all models contribute to the ensemble's success. Although embedding responses, clustering, and validation-based selection introduce some overhead, the overall method remains efficient and does not require additional training or fine-tuning.
For scenarios where computational efficiency is prioritized over maximum accuracy, we provide alternative configurations in Appendix A that achieve a balance between performance and model count.

\subsection{Discussion}
\label{sec:discussion}

Our extensive experimental results provide several key insights into the effectiveness and practical implications of DFPE. We organize our discussion around three main aspects: performance characteristics, Model Pool Homogeneity, parameter insensitivity, and trade-offs for practical considerations.
\paragraph{Performance Characteristics}
Our method’s strong results can be attributed to effectively leveraging a diverse set of models via clustering, quantile-based filtering, and adaptive weighting. By preserving model heterogeneity, DFPE capitalizes on complementary strengths, often retaining a majority of models for each subject while assigning higher influence to those that demonstrate subject-specific expertise. 
In addition, recent work has highlighted new ensembles like Mixtral-of-Experts (8x7B) from Mistral.ai\footnote{\url{https://mistral.ai/news/mixtral-of-experts/}} that achieve strong performance across multitask benchmarks. 
While some novel LLMs can surpass DFPE in absolute accuracy, they generally demand far larger computational resources. Notably, DFPE still exceeds certain significantly larger models, including LLaMA-2~70B, reinforcing the effectiveness of its ensemble approach. Incorporating those newer, larger LLMs into the DFPE model pool represents a promising direction for future work, potentially unlocking even further gains in performance through expanded ensemble diversity.

\paragraph{Model Pool Homogeneity}
It is important to note that the method's effectiveness hinges on the diversity present within the pool of base LLMs. Designed to capitalize on complementary strengths and varied perspectives from heterogeneous models, its benefits could diminish with a more homogeneous pool – for instance, models with similar architectures, training data, or biases. In such cases, fingerprint clustering may become less informative, and the ensemble would mainly function as averaging largely redundant model outputs. Consequently, performance gains achieved over a strong single model might be less pronounced, as the ensemble would lack the advantage of integrating genuinely diverse and contrasting insights. Future investigations could explore the method's performance under varying degrees of model pool homogeneity to further delineate its applicability and robustness.

\paragraph{Parameter Sensitivity and Trade-offs}
Our sensitivity analyses, shown in Figure~\ref{fig:sensitivity_analysis}, reveal important trade-offs between accuracy, computational cost, and model participation:

\textbf{Quantile Threshold}: Low thresholds (e.g., 0.05) maximize accuracy by including more models, increasing cost. Higher thresholds (0.15-0.20) offer a balanced approach, reducing ensemble size while maintaining performance.

\textbf{AccuracyFactor}: Higher values (around 5.0) prioritize strong models, but can risk over-reliance and reduce the benefits of ensemble diversity.

\textbf{DBSCAN Epsilon}: Small values create finer clusters, preserving subtle model differences. Large values lead to aggressive grouping and smaller ensembles.

For practical applications, we identify three deployment configurations that balance performance and efficiency:

\textbf{High-Accuracy Mode}: Uses most available models, achieving maximum accuracy at higher computational cost. Recommended for applications where accuracy is paramount.
    
\textbf{Balanced Mode}: Employs higher quantile thresholds and larger clustering epsilon, reducing ensemble size to 5-7 models while maintaining performance within 0.5-1.0\% of optimal.
    
\textbf{Efficient Mode}: Further reduces model count through more aggressive filtering, suitable for resource-constrained environments.

These configurations, detailed in Appendix A, provide practitioners with flexible options based on their specific requirements.

\section{Conclusion}
\label{sec:conclusion}

In this paper, we introduced DFPE, an ensemble method that leverages diversity and adaptivity to improve LLMs performance on multitask language understanding tasks. DFPE clusters models via “fingerprint” patterns, filters underperformers with a quantile-based threshold, and applies exponential weighting that emphasizes top-performing models while preserving valuable secondary perspectives. On the MMLU benchmark, DFPE achieves approximately 3\% higher overall accuracy and a 5\% boost in discipline-level accuracy over the best single model. These results underscore the importance of diverse solution strategies, selective filtering, and dynamic weighting. While DFPE currently focuses on multiple-choice tasks, applying its fingerprinting strategy to open-ended settings remains a promising direction. Future work also includes refining question-level adaptivity, exploring more scalable clustering, and addressing the unique challenges posed by large LLM pools. By balancing diversity, adaptivity, and efficiency, DFPE offers a practical and robust framework for high-performing ensembles across varied tasks. 

\section*{Limitations}
While DFPE demonstrates significant improvements, several limitations warrant discussion: \textbf{Dependence on Few-Shot Validation Data} DFPE relies on few-shot validation data for initial model selection and weighting. \textbf{Computational Overhead:} Running multiple models requires more computational resources than using a single small model. \textbf{Model Diversity Requirement:} The approach benefits from having a diverse pool of models with complementary strengths, and its effectiveness may reduced with homogeneous model sets;  \textbf{Task-Specificity:} The current implementation is confined to multiple-choice question answering.

\section*{Acknowledgements}
We used ChatGPT-4o for editing the language and refining the presentation of the text in this paper. The authors affirm that all research content and ideas are their own, and they take full responsibility for the final submitted manuscript.

\bibliography{acl_latex.bib}

\begin{thebibliography}{26}
\providecommand{\natexlab}[1]{#1}

\bibitem[{Abdin et~al.(2024)Abdin, Aneja, Awadalla, Awadallah, Awan, Bach, Bahree, Bakhtiari, Bao, Behl et~al.}]{abdin2024phi}
Marah Abdin, Jyoti Aneja, Hany Awadalla, Ahmed Awadallah, Ammar~Ahmad Awan, Nguyen Bach, Amit Bahree, Arash Bakhtiari, Jianmin Bao, Harkirat Behl, et~al. 2024.
\newblock Phi-3 technical report: A highly capable language model locally on your phone.
\newblock \emph{arXiv preprint arXiv:2404.14219}.

\bibitem[{Chang et~al.(2024)Chang, Wang, Wang, Wu, Yang, Zhu, Chen, Yi, Wang, Wang et~al.}]{chang2024survey}
Yupeng Chang, Xu~Wang, Jindong Wang, Yuan Wu, Linyi Yang, Kaijie Zhu, Hao Chen, Xiaoyuan Yi, Cunxiang Wang, Yidong Wang, et~al. 2024.
\newblock A survey on evaluation of large language models.
\newblock \emph{ACM Transactions on Intelligent Systems and Technology}, 15(3):1--45.

\bibitem[{Dubey et~al.(2024)Dubey, Jauhri, Pandey, Kadian, Al-Dahle, Letman, Mathur, Schelten, Yang, Fan et~al.}]{dubey2024llama}
Abhimanyu Dubey, Abhinav Jauhri, Abhinav Pandey, Abhishek Kadian, Ahmad Al-Dahle, Aiesha Letman, Akhil Mathur, Alan Schelten, Amy Yang, Angela Fan, et~al. 2024.
\newblock The llama 3 herd of models.
\newblock \emph{arXiv preprint arXiv:2407.21783}.

\bibitem[{Ester et~al.(1996)Ester, Kriegel, Sander, Xu et~al.}]{ester1996density}
Martin Ester, Hans-Peter Kriegel, J{\"o}rg Sander, Xiaowei Xu, et~al. 1996.
\newblock A density-based algorithm for discovering clusters in large spatial databases with noise.
\newblock In \emph{kdd}, volume~96, pages 226--231.

\bibitem[{Face(2023)}]{huggingface2023sentence}
Hugging Face. 2023.
\newblock Sentence transformers all-minilm-l6-v2.
\newblock \url{https://huggingface.co/sentence-transformers/all-MiniLM-L6-v2}.

\bibitem[{GLM et~al.(2024)GLM, Zeng, Xu, Wang, Zhang, Yin, Zhang, Rojas, Feng, Zhao et~al.}]{glm2024chatglm}
Team GLM, Aohan Zeng, Bin Xu, Bowen Wang, Chenhui Zhang, Da~Yin, Dan Zhang, Diego Rojas, Guanyu Feng, Hanlin Zhao, et~al. 2024.
\newblock Chatglm: A family of large language models from glm-130b to glm-4 all tools.
\newblock \emph{arXiv preprint arXiv:2406.12793}.

\bibitem[{Hendrycks et~al.(2020)Hendrycks, Burns, Basart, Zou, Mazeika, Song, and Steinhardt}]{hendrycks2020measuring}
Dan Hendrycks, Collin Burns, Steven Basart, Andy Zou, Mantas Mazeika, Dawn Song, and Jacob Steinhardt. 2020.
\newblock Measuring massive multitask language understanding.
\newblock \emph{arXiv preprint arXiv:2009.03300}.

\bibitem[{Huang et~al.(2024)Huang, Feng, Li, Xiang, Wang, Liu, and Qin}]{huang2024ensemble}
Yichong Huang, Xiaocheng Feng, Baohang Li, Yang Xiang, Hui Wang, Ting Liu, and Bing Qin. 2024.
\newblock Ensemble learning for heterogeneous large language models with deep parallel collaboration.
\newblock In \emph{The Thirty-eighth Annual Conference on Neural Information Processing Systems}.

\bibitem[{Jiang et~al.(2023{\natexlab{a}})Jiang, Sablayrolles, Mensch, Bamford, Chaplot, Casas, Bressand, Lengyel, Lample, Saulnier et~al.}]{jiang2023mistral}
Albert~Q Jiang, Alexandre Sablayrolles, Arthur Mensch, Chris Bamford, Devendra~Singh Chaplot, Diego de~las Casas, Florian Bressand, Gianna Lengyel, Guillaume Lample, Lucile Saulnier, et~al. 2023{\natexlab{a}}.
\newblock Mistral 7b.
\newblock \emph{arXiv preprint arXiv:2310.06825}.

\bibitem[{Jiang et~al.(2023{\natexlab{b}})Jiang, Ren, and Lin}]{jiang2023llm}
Dongfu Jiang, Xiang Ren, and Bill~Yuchen Lin. 2023{\natexlab{b}}.
\newblock Llm-blender: Ensembling large language models with pairwise ranking and generative fusion.
\newblock \emph{arXiv preprint arXiv:2306.02561}.

\bibitem[{Jiang et~al.(2023{\natexlab{c}})Jiang, Ruan, Huang, Liao, Pitis, Grosse, and Ba}]{jiang2023calibrating}
Mingjian Jiang, Yangjun Ruan, Sicong Huang, Saifei Liao, Silviu Pitis, Roger~Baker Grosse, and Jimmy Ba. 2023{\natexlab{c}}.
\newblock Calibrating language models via augmented prompt ensembles.

\bibitem[{Lu et~al.(2024)Lu, Pang, Xiao, Zhu, Xia, and Zhang}]{lu2024merge}
Jinliang Lu, Ziliang Pang, Min Xiao, Yaochen Zhu, Rui Xia, and Jiajun Zhang. 2024.
\newblock Merge, ensemble, and cooperate! a survey on collaborative strategies in the era of large language models.
\newblock \emph{arXiv preprint arXiv:2407.06089}.

\bibitem[{Lu et~al.(2023)Lu, Yuan, Lin, Lin, Yuan, Zhou, and Zhou}]{lu2023routing}
Keming Lu, Hongyi Yuan, Runji Lin, Junyang Lin, Zheng Yuan, Chang Zhou, and Jingren Zhou. 2023.
\newblock Routing to the expert: Efficient reward-guided ensemble of large language models.
\newblock \emph{arXiv preprint arXiv:2311.08692}.

\bibitem[{Matarazzo and Torlone(2025)}]{matarazzo2025survey}
Andrea Matarazzo and Riccardo Torlone. 2025.
\newblock A survey on large language models with some insights on their capabilities and limitations.
\newblock \emph{arXiv preprint arXiv:2501.04040}.

\bibitem[{Maurya et~al.(2024)Maurya, Srivatsa, and Kochmar}]{maurya2024selectllm}
Kaushal~Kumar Maurya, KV~Srivatsa, and Ekaterina Kochmar. 2024.
\newblock Selectllm: Query-aware efficient selection algorithm for large language models.
\newblock \emph{arXiv preprint arXiv:2408.08545}.

\bibitem[{Mavromatis et~al.(2024)Mavromatis, Karypis, and Karypis}]{mavromatis2024pack}
Costas Mavromatis, Petros Karypis, and George Karypis. 2024.
\newblock Pack of llms: Model fusion at test-time via perplexity optimization.
\newblock \emph{arXiv preprint arXiv:2404.11531}.

\bibitem[{Pitis et~al.(2023)Pitis, Zhang, Wang, and Ba}]{pitis2023boosted}
Silviu Pitis, Michael~R Zhang, Andrew Wang, and Jimmy Ba. 2023.
\newblock Boosted prompt ensembles for large language models.
\newblock \emph{arXiv e-prints}, pages arXiv--2304.

\bibitem[{Team et~al.(2024)Team, Mesnard, Hardin, Dadashi, Bhupatiraju, Pathak, Sifre, Rivi{\`e}re, Kale, Love et~al.}]{gemma_2024}
Gemma Team, Thomas Mesnard, Cassidy Hardin, Robert Dadashi, Surya Bhupatiraju, Shreya Pathak, Laurent Sifre, Morgane Rivi{\`e}re, Mihir~Sanjay Kale, Juliette Love, et~al. 2024.
\newblock Gemma: Open models based on gemini research and technology.
\newblock \emph{arXiv preprint arXiv:2403.08295}.

\bibitem[{Tekin et~al.(2024)Tekin, Ilhan, Huang, Hu, and Liu}]{tekin2024llm}
Selim Tekin, Fatih Ilhan, Tiansheng Huang, Sihao Hu, and Ling Liu. 2024.
\newblock Llm-topla: Efficient llm ensemble by maximising diversity.
\newblock In \emph{Findings of the Association for Computational Linguistics: EMNLP 2024}, pages 11951--11966.

\bibitem[{Wang et~al.(2023)Wang, Aitchison, and Rudolph}]{wang2023lora}
Xi~Wang, Laurence Aitchison, and Maja Rudolph. 2023.
\newblock Lora ensembles for large language model fine-tuning.
\newblock \emph{arXiv preprint arXiv:2310.00035}.

\bibitem[{Xu et~al.(2025)Xu, Chen, Wu, and Zhang}]{xu2025hit}
Yangyifan Xu, Jianghao Chen, Junhong Wu, and Jiajun Zhang. 2025.
\newblock Hit the sweet spot! span-level ensemble for large language models.
\newblock In \emph{Proceedings of the 31st International Conference on Computational Linguistics}, pages 8314--8325.

\bibitem[{Xu et~al.(2024)Xu, Lu, and Zhang}]{xu2024bridging}
Yangyifan Xu, Jinliang Lu, and Jiajun Zhang. 2024.
\newblock Bridging the gap between different vocabularies for llm ensemble.
\newblock In \emph{Proceedings of the 2024 Conference of the North American Chapter of the Association for Computational Linguistics: Human Language Technologies (Volume 1: Long Papers)}, pages 7133--7145.

\bibitem[{Yang et~al.(2024)Yang, Yang, Zhang, Hui, Zheng, Yu, Li, Liu, Huang, Wei et~al.}]{yang2024qwen2}
An~Yang, Baosong Yang, Beichen Zhang, Binyuan Hui, Bo~Zheng, Bowen Yu, Chengyuan Li, Dayiheng Liu, Fei Huang, Haoran Wei, et~al. 2024.
\newblock Qwen2. 5 technical report.
\newblock \emph{arXiv preprint arXiv:2412.15115}.

\bibitem[{Young et~al.(2024)Young, Chen, Li, Huang, Zhang, Zhang, Wang, Li, Zhu, Chen et~al.}]{young2024yi}
Alex Young, Bei Chen, Chao Li, Chengen Huang, Ge~Zhang, Guanwei Zhang, Guoyin Wang, Heng Li, Jiangcheng Zhu, Jianqun Chen, et~al. 2024.
\newblock Yi: Open foundation models by 01. ai.
\newblock \emph{arXiv preprint arXiv:2403.04652}.

\bibitem[{Zheng et~al.(2024)Zheng, Wang, Liang, Chen, Zheng, and Wang}]{zheng2024efficiently}
Guorui Zheng, Xidong Wang, Juhao Liang, Nuo Chen, Yuping Zheng, and Benyou Wang. 2024.
\newblock Efficiently democratizing medical llms for 50 languages via a mixture of language family experts.
\newblock \emph{arXiv preprint arXiv:2410.10626}.

\bibitem[{Zhu et~al.(2024)Zhu, Frick, Wu, Zhu, Ganesan, Chiang, Zhang, and Jiao}]{starling2023}
Banghua Zhu, Evan Frick, Tianhao Wu, Hanlin Zhu, Karthik Ganesan, Wei-Lin Chiang, Jian Zhang, and Jiantao Jiao. 2024.
\newblock Starling-7b: Improving helpfulness and harmlessness with rlaif.
\newblock In \emph{First Conference on Language Modeling}.

\end{thebibliography}

\clearpage

\appendix
\section{Appendix A}

While our main results focus on the optimal accuracy configuration, practitioners often need to balance performance gains against computational costs. Here we present a detailed analysis of an alternative configuration that achieves near-optimal performance while significantly reducing computational overhead.

\subsection{Balanced Configuration Parameters}
We identified a balanced configuration with the following parameters:
\begin{itemize}[leftmargin=*,noitemsep]
    \item Quantile threshold: 0.5 (vs. 0.05 in optimal setting)
    \item AccuracyFactor: 7 (vs. 5 in optimal setting)
    \item DBSCAN Epsilon: 0.001 (vs. 0.0001 in optimal setting)
\end{itemize}

This configuration achieves an average accuracy of 72.5\% (1\% below optimal) while reducing the mean number of models per subject to 6 (compared to 9 in the optimal setting).

\subsection{Model Selection Analysis}

\begin{figure}[h]
\centering
\includegraphics[width=\linewidth]{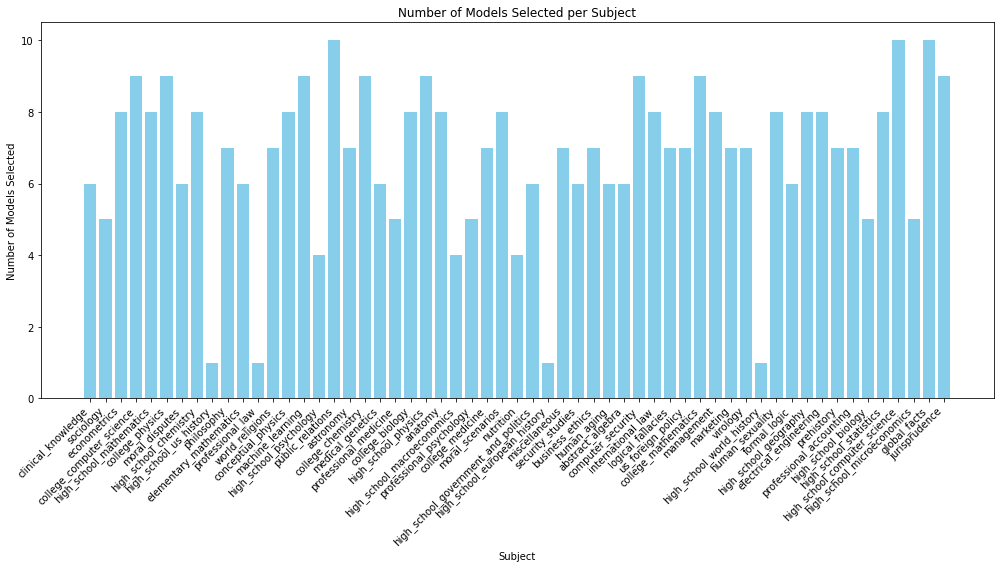}
\caption{Distribution of selected models per subject. The variation in bar heights demonstrates how DFPE adapts its ensemble size to subject-specific requirements while maintaining efficiency.}
\label{fig:Models_Per_Sub}
\end{figure}

Figure~\ref{fig:Models_Per_Sub} shows the distribution of selected models across subjects. Several key patterns emerge:
\begin{itemize}[leftmargin=*,noitemsep]
    \item Variation in model count: The number of selected models varies from 1 to 10 across subjects
    \item Subject-specific adaptation: Different subjects benefit from different ensemble sizes
    \item Consistent core: Most subjects maintain 6-8 models, suggesting a natural balance point
\end{itemize}

\subsection{Model Co-occurrence Analysis}

\begin{figure}[h]
\centering
\includegraphics[width=\linewidth]{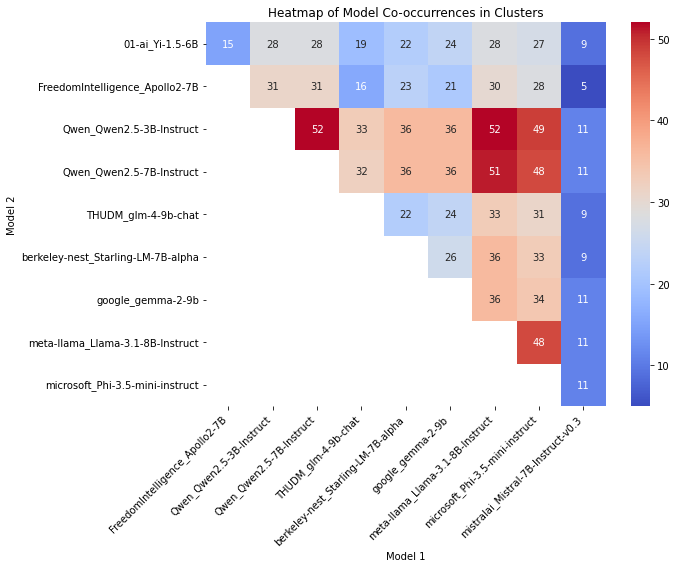}
\caption{Heatmap of model co-occurrences within clusters. Cell values indicate frequency of model pairs being selected together. The diagonal is zero by definition. Higher values (darker colors) suggest stronger complementarity between models.}
\label{fig:model_cooccurrence}
\end{figure}

To understand model relationships, we analyzed their co-occurrence patterns within clusters (Figure~\ref{fig:model_cooccurrence}). Our analysis reveals several interesting patterns:

\begin{itemize}[leftmargin=*,noitemsep]
    \item \textbf{Strong Partnerships}:
        - Qwen family models (Qwen2.5-3B and Qwen2.5-7B-Instruct) show highest co-occurrence (52 instances)
        - Strong affinity between Qwen models and Llama-3.1-8B (51-52 co-occurrences)
        - Phi-3.5-mini frequently pairs with Qwen models (48-49 instances)
    
    \item \textbf{Complementary Groups}:
        - Models cluster into "specialists" and "generalists"
        - Lower co-occurrence patterns indicate complementary strengths
    
    \item \textbf{Model Independence}:
        - Some models show consistent independence
        - Suggests unique capabilities or specialization
\end{itemize}

\subsection{Practical Implications}

These findings have several important implications for practitioners:

\textbf{Resource Optimization}: The balanced configuration offers a practical trade-off between performance and computational cost

\textbf{Model Selection}: Strong co-occurrence patterns can guide initial model selection when building new ensembles

\textbf{Deployment Strategy}: Subject-specific ensemble sizes suggest opportunities for dynamic resource allocation

\end{document}